\def\cvprPaperID{5980}
\def\confName{CVPR}
\def\confYear{2023}

\def\paperTitle{PanelNet: Understanding 360 Indoor Environment via Panel Representation}

\def\authorBlock{
    Haozheng Yu \qquad
    Lu He \qquad
    Bing Jian \qquad
    Weiwei Feng \qquad
    Shan Liu
    \\
    Tencent America \\
   {\tt\small \{haozhengyu, lhluhe, bingjian, wfeng, shanl\}@tencent.com}
   }

\newif\ifreview 
\newif\ifarxiv \newcommand{\arxiv}{\arxivtrue}
\newif\ifcamera 
\newif\ifrebuttal 

\arxiv 

\pdfoutput=1
\documentclass[10pt,twocolumn,letterpaper]{article}
\ifreview \usepackage[review]{cvpr} \fi
\ifarxiv \usepackage[pagenumbers]{cvpr} \fi
\ifrebuttal \usepackage[rebuttal]{cvpr} \fi
\ifcamera \usepackage{cvpr} \fi

\usepackage{graphicx}
\usepackage{amsmath}
\usepackage{amssymb}
\usepackage{booktabs}

\usepackage{times}
\usepackage{microtype}
\usepackage{epsfig}
\usepackage[table,xcdraw]{xcolor}
\usepackage{caption}
\usepackage{float}
\usepackage{placeins}
\usepackage{color, colortbl}
\usepackage{stfloats}
\usepackage{enumitem}
\usepackage{tabularx}
\usepackage{xstring}
\usepackage{multirow}
\usepackage{xspace}
\usepackage{url}
\usepackage{subcaption}
\usepackage{xcolor}
\usepackage[hang,flushmargin]{footmisc}

\ifcamera \usepackage[accsupp]{axessibility} \fi

% ...

\ifarxiv  \fi

\newcommand{\R}[1]{{%
    \textbf{%
        \ifstrequal{#1}{1}{\textcolor{red}{R#1}}{%
        \ifstrequal{#1}{2}{\textcolor{blue}{R#1}}{%
        \ifstrequal{#1}{3}{\textcolor{magenta}{R#1}}{%
        \ifstrequal{#1}{4}{\textcolor{teal}{R#1}}{%
                           \textcolor{cyan}{R#1}%
        }}}}%
    }%
}}

\usepackage{xr-hyper}

\makeatletter
\newcommand*{\addFileDependency}[1]{
  \typeout{(#1)}
  \@addtofilelist{#1}
  \IfFileExists{#1}{}{\typeout{No file #1.}}
}

\makeatother

\usepackage[pagebackref,breaklinks,colorlinks]{hyperref}
\usepackage[capitalize]{cleveref}
\crefname{section}{Sec.}{Secs.}
\crefname{table}{Table}{Tables}
\crefname{figure}{Fig.}{Figs.}

\begin{document}
\title{\paperTitle}
\author{\authorBlock}
\maketitle

\begin{abstract}

Indoor 360 panoramas have two essential properties. (1) The panoramas are continuous and seamless in the horizontal direction. (2) Gravity plays an important role in indoor environment design. By leveraging these properties, we present PanelNet, a framework that understands indoor environments using a novel panel representation of 360 images. We represent an equirectangular projection (ERP) as consecutive vertical panels with corresponding 3D panel geometry. To reduce the negative impact of panoramic distortion, we incorporate a panel geometry embedding network that encodes both the local and global geometric features of a panel. To capture the geometric context in room design, we introduce Local2Global Transformer, which aggregates local information within a panel and panel-wise global context. It greatly improves the model performance with low training overhead. Our method outperforms existing methods on indoor 360 depth estimation and shows competitive results against state-of-the-art approaches on the task of indoor layout estimation and semantic segmentation.

\end{abstract}
\section{Introduction}
\label{sec:intro}

Understanding indoor environments is an important topic in computer vision as it is crucial for multiple practical applications such as room reconstruction, robot navigation, and virtual reality applications. Early methods focus on modeling indoor scenes using perspective images~\cite{eigen2014depth, eigen2015predicting, laina2016FCRN}. 
With the development of CNNs and omnidirectional photography, many works turn to understand indoor scenes using panorama images. Compared to the perspective images, panorama images have a larger field-of-view (FoV)~\cite{zhang2014panocontext} and provide the geometric context of the indoor environment in a continuous way~\cite{Pintore2021Slicenet}. 

There are several 360 input formats used in indoor scene understanding. One of the most commonly-used formats is the equirectangular projection (ERP). Modeling the holistic scene from an ERP is challenging. The ERP distortion increases when pixels are close to the zenith or nadir of the image, which may decrease the power of the convolutional structures designed for distortion-free perspective images. To eliminate the negative effects of ERP distortion, recent works~\cite{eder2020tangent, li2022omnifusion, rey2022360monodepth} focus on 
decomposing the whole panorama into perspective patches, i.e., tangent images. 
However, partitioning a panorama into discontinuous patches
breaks the local continuity of gravity-aligned scenes and objects
which limits the performance of these works. To reduce the impact of distortion while preserving the local continuity, we present PanelNet, a novel network to understand the indoor scene from equirectangular projection.

\begin{figure}[t!]
 \centering
 \includegraphics[width=0.5\textwidth]{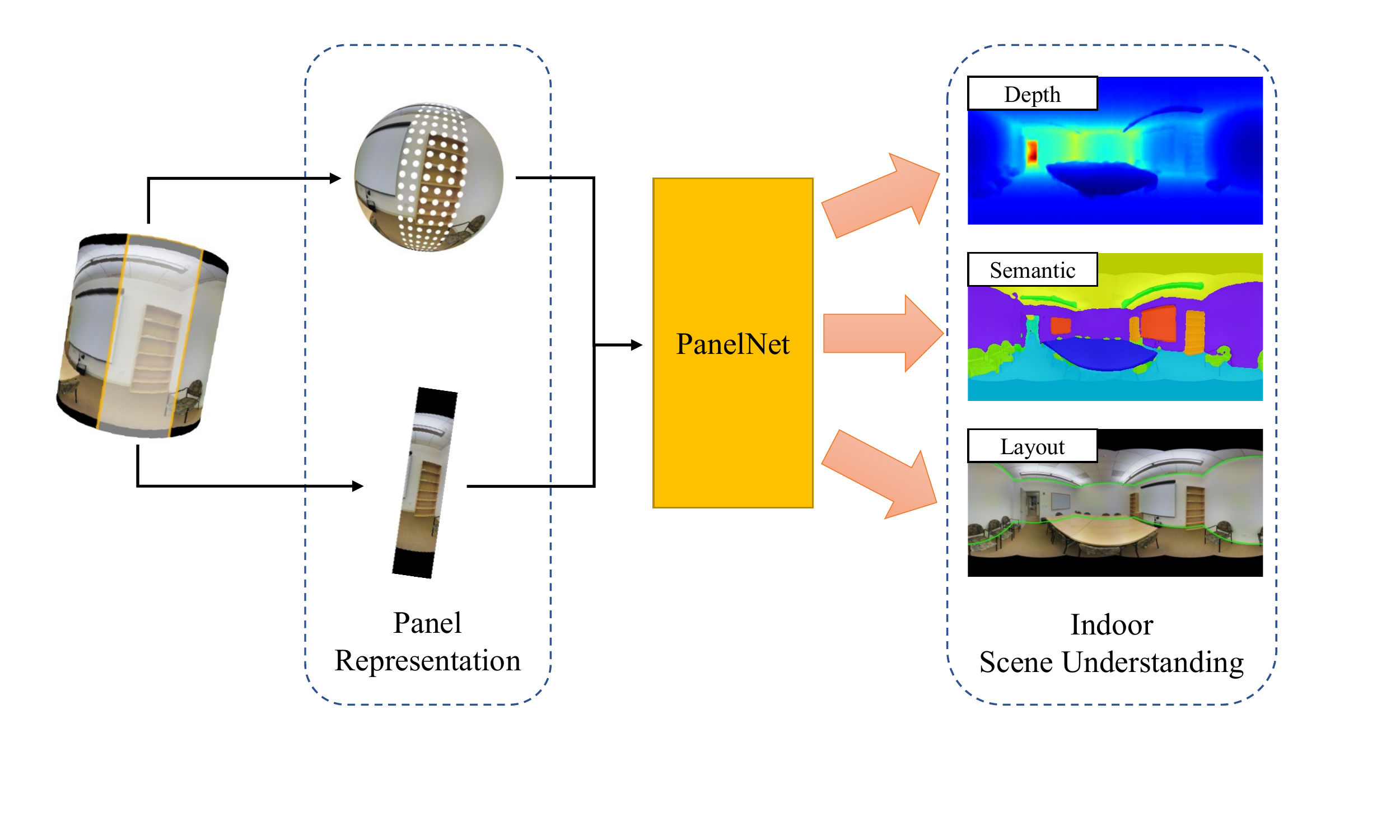}
 \vspace{-10mm}
 \caption{An overview of the proposed system. We present PanelNet, a network that learns the indoor environment using a novel panel representation of ERP. We formulate the panel representation as consecutive ERP panels with corresponding global and local geometry. By slightly modifying the network structure, PanelNet is capable of tackling major 360 indoor understanding tasks such as depth estimation, semantic segmentation and layout prediction.}
 \vspace{-3mm}
 \label{fig:diagrame}
\end{figure}

We design our PanelNet based on two essential properties of equirectangular projection. (1) The ERP is continuous and seamless in the horizontal direction. (2) Gravity plays an important role in indoor environment design, which makes the gravity-aligned features crucial for indoor 360 understanding~\cite{SunSC21, Pintore2021Slicenet}. Following these two properties, we tackle the challenges above through a novel panel representation of ERP. We represent an ERP as consecutive panels with corresponding global and local 3D geometry, which preserves the gravity-aligned features within a panel and maintains the global continuity of the indoor structure across panels. Inspired by Omnifusion~\cite{li2022omnifusion}, we design a geometry embedding network for panel representations that encodes both local and global features of panels to reduce the negative effects of ERP distortion without adding further explicit distortion fixing modules. We further introduce Local2Global Transformer as a feature processor. Considering the nature of panel representation, we design this Transformer with Window Blocks for local information aggregation and Panel Blocks for panel-wise context capturing. The main contributions of our work are: 
\begin{itemize} 
    \setlength{\itemsep}{0pt}
    \item We represent the ERP as consecutive vertical panels with corresponding 3D geometry. We introduce PanelNet, a novel indoor panorama understanding pipeline using the panel representation. Following the essential geometric properties of the indoor equirectangular projection, our framework outperforms existing methods on the task of indoor 360 depth estimation and shows competitive results on other indoor scene understanding tasks such as semantic segmentation and layout prediction.
    \item We propose a panel geometry embedding network that encodes both local and global geometric features of panels and reduces the negative impact of ERP distortion implicitly while preserving the geometric continuity.
    \item We design Local2Global Transformer as a feature processor, which greatly enhances the continuity of geometric features and improves the model performance by successfully aggregating the local information within a panel and capturing panel-wise context accurately. 


\end{itemize}
\section{Related Work}
\label{sec:related}

We aim to design a general framework to tackle the major tasks of indoor scene understanding using 360 images. We briefly review the related works.

\subsection{Monocular depth estimation}
Estimating the depth from an image is an essential problem in computer vision. Early works tackle this problem via stereo matching~\cite{scharstein2002taxonomy} and motion clues in a video~\cite{karsch2014depth}. With the flourishment of deep learning, researchers develop monocular depth estimation methods via deep neural networks. Eigen et al.~\cite{eigen2014depth} first develop a multi-scale deep neural network to regress depth from a single image. Their later work~\cite{eigen2015predicting} introduces a more general multi-scale network with a VGG encoder for predicting depth, surface normal and semantic labels. Laina et al.~\cite{laina2016FCRN} design a fully convolutional residual network with upsampling layers. They also introduce Berhu Loss for network training.  Cao et al.~\cite{cao2017estimating} formulate the depth regression task as a classification task and apply fully-connected Conditional Random Fields (CRF) to obtain the final depth prediction. Other works address this problem with different strategies. Fu et al.~\cite{fu2018deep} introduce dilated convolutions to enlarge receptive fields and utilize an ordinary regression loss for network optimization. Geometric constraints are also often exploited to enhance potential geometry relationships~\cite{qi2018geonet, yin2019enforcing}.

\subsection{Panorama depth estimation}
One key limitation for understanding the scene via a perspective image is the lack of geometric context due to the small FoV. Recently, the development of 360 imagery and the popularity of 360 cameras encourage researchers to address the scene understanding tasks directly on panoramas. Compared to perspective images, panoramas preserve the structural context of the room while introducing distortion. Recent works estimate the depth from panoramas by jointly learning the room structure~\cite{zeng2020joint, jin2020geometric}, planes and normals~\cite{eder2019pano, He_2022_CVPR}. By leveraging the gravity-aligned features in indoor panoramas, Pintore et al.~\cite{pintore2020atlantanet} and Sun et al.~\cite{SunSC21} design networks that directly work on the equirectangular projections. However, directly applying convolution-based structures designed for distortion-free perspective images on panoramas may lead to sub-optimal results~\cite{zioulis2018omnidepth}. To reduce the negative impact of panorama distortion, several works design distortion-aware CNNs~\cite{coors2018spherenet, tateno2018distortion, eder2019mapped, zhuang2022acdnet} based on the nature of ERP distortion. Other methods handle this problem via less-distortion representations instead of directly modeling the distortion. Wang et al.~\cite{BiFuse20} and Jiang et al.~\cite{jiang2021unifuse} fuse the cubemap projection with ERP to mimic peripheral and foveal vision as the human eye. Recently, Eder et al.~\cite{eder2020tangent} propose to handle panoramic distortion with tangent representation, which inspires further studies on tangent-based depth estimation such as Omnifusion~\cite{li2022omnifusion} and 360MonoDepth~\cite{rey2022360monodepth}. However, these methods introduce discrepancies between the patches that are hard to be removed by the fusion modules. We tackle these challenges by introducing the panel representation of ERP. The panels are directly extracted from the original panorama via a sliding window mechanism similar to CNNs, which preserves the gravity-aligned information of indoor scenes within panels. Rather than fixing panoramic distortion in an explicit way, we add a panel geometry embedding network to learn the distortion for panels and reduce the negative impact of distortion with minimal computation cost.

\subsection{Other indoor understanding tasks}
360 semantic segmentation is another important dense prediction task for indoor understanding. Similar to 360 depth estimation, most of the recent panorama semantic segmentation works focus on reducing the negative impact of ERP distortion~\cite{tateno2018distortion, jiang2018spherical, lee2019spherephd}. Other approaches try different strategies such as joint learning the semantic labels with layout~\cite{zhang2021deeppanocontext} and unsupervised transfer learning~\cite{zhang2022bending}.

For layout estimation, previous approaches model this task as a dense prediction task. Zou et al.~\cite{zou2018layoutnet} design a network in U-Net structure to jointly learn the layout boundaries and corner positions from the input RGB image and Manhattan line map. Yang et al.~\cite{yang2019dula} introduce a two-branch network that incorporates both equirectangular projection and perspective ceiling view to learn different layout clues. Sun et al.~\cite{SunHorizon19} simplify the layout estimation task from dense prediction to three 1-D boundary predictions. They also propose a panorama stretch method that can diversify the panorama data as data augmentation. Wang et al.~\cite{wang2021led} transform layout estimation to depth prediction on the horizon line of a panorama. They  design a layout-to-depth transformation to convert the layout into horizon-depth via ray casting. Jiang et al.~\cite{jiang2022lgt} represent the room layout as the floor boundary and height. We follow this representation when predicting room layouts using our modified PanelNet. 

\begin{figure*}[!t]
  \centering
    \includegraphics[width=1\linewidth]{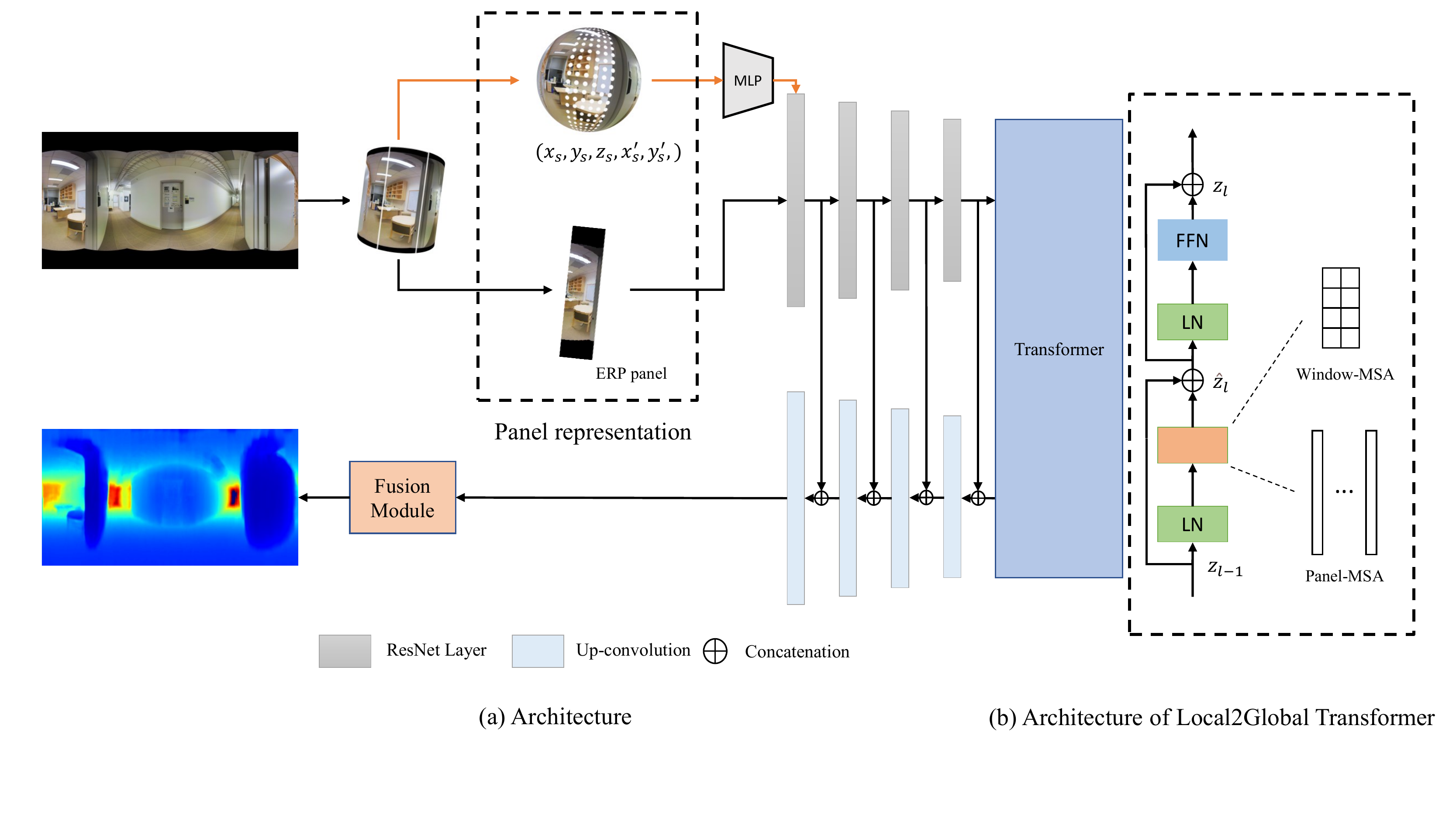}
    \vspace{-15mm}
    \caption{The architecture of the network. Given the stride and interval, an ERP is first partitioned into consecutive panels by a sliding window. Meanwhile, for each pixel on every panel, the corresponding global coordinates are represented as its absolute 3D coordinates $(x_s, y_s, z_s)$. Its local coordinates are represented as its relative 3D coordinates to the panel $(x^\prime_s, y^\prime_s)$. The local and global coordinates are used as input of an MLP to generate geometric features. The Local2Global Transformer is applied to aggregate local information (Window Blocks) and capture global dependencies (Panel Blocks). In each Transformer block, we stack LayerNorm (LN), multi-head self-attention module (W/P-MSA) and Forward Feed Network (FFN) with skip connection as shown in (b).
  } 
  \vspace{-1mm}
  \label{fig:system}
\end{figure*}

\subsection{Vision Transformer}
Transformers are originally proposed in the field of natural language processing~\cite{vaswani2017attention} and soon become very popular due to their superior performance on NLP tasks. ViT~\cite{dosovitskiy2020image} and its following works~\cite{liu2021swin, yuan2021tokens, ranftl2021vision, liu2022swinv2} demonstrate that Transformers are suitable for capturing long-range dependencies for vision tasks by achieving superior results against CNN based models on image classification, image segmentation and dense prediction. Transformers are also used for 360 indoor understanding tasks such as depth estimation~\cite{li2022omnifusion}, layout estimation~\cite{jiang2022lgt} and semantic segmentation~\cite{zhang2022bending}. We design a Local2Global Transformer as a feature processor, which contains Window Blocks to aggregate local information within a panel and Panel Blocks to capture the long-range relationships among the panels.
Our proposed Local2Global Transformer greatly improves the model performance on 360 indoor understanding tasks.


\section{Method}
\label{sec:method}


\subsection{Network architecture} \label{Network architecture}
As illustrated in Figure~\ref{fig:system}, we implement our network in an encoder-decoder fashion. We incorporate a panel geometry embedding network to reduce the negative impact of panorama distortion and Local2Global Transformer to aggregate local and global information.

\noindent\textbf{Panel Representation of ERP } Given the stride $S$ and the interval $I$ of the panels, an input RGB ERP in resolution $H_e \times W_e$ is divided into $N$ consecutive panels by a vertical sliding window in size $H_e \times I$. Since an ERP is continuous and seamless in the horizontal direction, we extract the panels across the left and right edges of the ERP. So $N = \frac{W_e}{S}$. The corresponding local and global geometric features are generated together given $I$ and $S$, details discussed in Section \ref{Geometry embedding}.

\noindent\textbf{Backbone } We use a ResNet-34~\cite{he2016deep} based architecture as the feature extractor of our model. It takes the ERP panels as input and generates the feature maps of each panel in 4 different scales. We apply a $1 \times 1$ convolution layer to reduce the dimensions of the final feature map of each panel to $f_b \in \mathbb{R}^ {C_b \times H_b \times W_b}$, where $H_b = \frac{H_e}{32}$, $W_b = \frac{I}{32}$, $C_b = \frac{D}{H_b \times W_b}$ and $D = 512 $ for any interval and stride. The feature maps are then used as input of the Local2Global Transformer for information aggregation, discussed in Section~\ref{Transformer}.



\noindent\textbf{Decoder } As illustrated in Figure~\ref{fig:system}. For each decoder layer, we concatenate the feature map with the feature map generated by the corresponding encoder layer. We apply up-convolutions to gradually recover the feature map to the RGB input resolution. Similar to Omnifusion~\cite{li2022omnifusion}, we predict a learnable confidence map by the decoder to improve the final merging result. For the final merge, we take the average of the prediction of all panels. By slightly modifying the network structure, our model is capable of other indoor 360 dense prediction tasks such as semantic segmentation. For 360 layout estimation, we follow the layout representation of LGT-Net~\cite{jiang2022lgt} and represent the room layouts as floor boundary and room height. We add one linear layer to generate floor boundary after the last decoder layer and two linear layers to generate room height. The default length of the output 1-D floor boundary is 1024.

\subsection{Local2Global Transformer} \label{Transformer}
Although partitioning the ERP into consecutive panels via a sliding window preserves the continuity of indoor structure, capturing the long-range dependencies is still crucial. Since the ERP is seamless in the horizontal direction, two distant panels on a panorama have a closer realistic distance. To address this problem and further improve local information aggregation, we present Local2Global Transformer, which consists of two major important components. Window Blocks to enhance the geometry relations within a local panel and Panel Blocks for capturing long-range context.


In Window Blocks, we compute the window multi-head self-attention similar to ViT~\cite{dosovitskiy2020image}. For each panel, we reshape the input feature map $f_b \in \mathbb{R}^ {C_b \times H_b \times W_b}$ into a sequence of flattened 2D feature patches $f_w \in \mathbb{R}^ {N_w \times (P^2 \cdot C_b)}$, where $(P \times P)$ is the size of the feature patch and $N_w=\frac{H_bW_b}{P^2}$ is the number of feature patches in current Window Block. In our experiment, $P=1,2,4$ for Window Blocks in different resolutions. Similar to ViT~\cite{dosovitskiy2020image}, we apply a learnable position embedding $E_w \in \mathbb{R}^ {N_w \times (P^2 \cdot C_b)}$ to maintain the positional information of feature patches. 

In Panel Blocks, we aim to aggregate global information via panel-wise multi-head self-attention. The feature maps of all panels are compressed to $N$ 1-D feature vectors $f_p \in \mathbb{R}^ {N \times D}$ and then used as tokens in the Panel Blocks. Similar to Window Blocks, we add a learnable positional embedding $E_p \in \mathbb{R}^ {N \times D}$ to the tokens to retain patch-wise positional information. See more discussion of positional embedding in Section~\ref{subSection: Ablation study}.

Following the standard Transformer block architecture of ViT~\cite{dosovitskiy2020image}, we stack multi-head self-attention module (MSA) and Feed-Forward Network (FFN). We apply a LayerNorm (LN) before each MSA and FFN. A Local2Global Transformer block is computed as

\begin{small}
\begin{equation}
        \begin{array}{l}
        \hat{z}_l = {\rm(W/P)\text{-}MSA} ({\rm LN}(z_{l-1})) + z_{l-1} \\
        z_l = {\rm FFN} ({\rm LN}(\hat{z}_l)) + \hat{z}_l \\
        \end{array}
\end{equation}
\end{small}


\noindent where $l$ is the block number of each stage. $\hat{z}_l$ and $z_l$ is the output feature map of the Window/Panel - MSA and FFN. To aggregate the features from local to global, we stack the Window Blocks according to the window size from small to high successively. The Panel Blocks are stacked after the Window Blocks. For the best performance, we use 12 Transformer blocks and place them in this order: Low-Res W-Blocks(2), Mid-Res W-Blocks(2), High-Res W-blocks(2), Panel Blocks(6). We observe a performance drop when shuffling this order because the compress operation in Panel Blocks reduces the impact of local information aggregation performed by Window Blocks.

\subsection{Panel geometry embedding} \label{Geometry embedding}

Inspired by the geometry fusion pipeline of Omnifusion~\cite{li2022omnifusion}, we develop a panel geometry embedding module to combine the geometry feature with the image feature together and reduce the negative impact of the ERP distortion. For a pixel $P_e(x_e, y_e)$ located on an ERP, the absolute 3D world position of its counterpart located on a unit sphere $P_s(\varphi, \theta)$ can be calculated as:



\begin{equation}
    \left\{
        \begin{array}{l}
        x_s = \sin \theta \cos \varphi \\
        y_s = \sin \theta \sin \varphi \\
        z_s = \cos \theta
        \end{array}
    \right.
\end{equation}

\noindent where $\varphi$ and $\theta$ are the azimuth angle and the polar angle of the point on a sphere, respectively. The 3D world coordinates $P_s(x_s, y_s, z_s)$ are then used to generate global features. Since each ERP panel has the same distortion, the relative position of each pixel to the panel where it is located is also important. Similar to the absolute 3D position, we assign a relative 3D local position $P_s(x^\prime_s, y^\prime_s, z^\prime_s)$ for each pixel per panel. We use the global 3D world coordinates of a randomly selected panel to represent the relative 3D position of all panels, which is unchanged during the entire experiment. Note that $z_s = z^\prime_s$. So the final input of a point on a panel to the geometry embedding network is the combination of its local and global coordinates $(x_s, y_s, z_s, x^\prime_s, y^\prime_s)$. 

We generate global and local geometric features via a two-layer MLP. The generated geometry features are added to the first layer of the backbone to make the network aware of ERP distortion.

\subsection{Loss function}
For depth estimation, we follow the previous works and train the network by minimizing the \textit{Reverse Huber Loss (BerHu)}~\cite{laina2016FCRN} in a fully supervised way.

\begin{equation}
    B(e) = 
    \left\{
        \begin{array}{lr}
        \left| e \right| & \left| e \right| \leq c,\\
        \frac{e^2+c^2}{2c} & \left| e \right| > c.
        \end{array}
    \right.
\end{equation}

\noindent where $e$ is the error term and the threshold $c$ determines where the switch from L1 to L2 occurs. For semantic segmentation, we use Cross-Entropy Loss with class-wise weights to balance the examples. For layout estimation, we strictly follow LGT-Net~\cite{jiang2022lgt} and use the combination of L1 loss for horizon depth and room height, normal loss and normal gradient loss to train our network.
\section{Experiments}
\label{sec:experiments}

\begin{table*}[!t]
  \centering
  \begin{tabular}{@{}cl||ccccccc@{}}
    \toprule
    Dataset & Method & MRE $\downarrow$ & MAE $\downarrow$ & RMSE $\downarrow$ & RMSE (log) $\downarrow$ & $\delta^1$ $\uparrow$ & $\delta^2$ $\uparrow$ & $\delta^3$ $\uparrow$ \\
    \midrule
    \multirow{7}{*}{Stanford2D3D} & FCRN~\cite{laina2016FCRN} & 0.1837 & 0.3428 & 0.5774 & 0.1100 & 0.7230 & 0.9207 & 0.9731 \\
     & OmniDepth ~\cite{zioulis2018omnidepth} & 0.1996 & 0.3743 & 0.6152 & 0.1212 & 0.6877 & 0.8891 & 0.9578 \\
     & Bifuse~\cite{BiFuse20} & 0.1209 & 0.2343 & 0.4142 & 0.0787 & 0.8660 & 0.9580 & 0.9860 \\
     & HoHoNet~\cite{SunSC21} & 0.1014 & 0.2027 & 0.3834 & 0.0668 & 0.9054 & 0.9693 & 0.9886 \\
     & SliceNet~\cite{Pintore2021Slicenet} & 0.1043 & 0.1838 & 0.3689 & 0.0771 & 0.9034 & 0.9645 & 0.9864 \\
     & Omnifusion~\cite{li2022omnifusion} & 0.1031 & 0.1958 & 0.3521 & 0.0698 & 0.8913 & 0.9702 & 0.9875  \\
     & Ours & \bf{0.0829} & \bf{0.1520} & \bf{0.2933} & \bf{0.0579} & \bf{0.9242} & \bf{0.9796} & \bf{0.9915} \\
    \midrule
    \multirow{7}{*}{Matterport3D} & FCRN~\cite{laina2016FCRN} & 0.2409 & 0.4008 & 0.6704 & 0.1244  & 0.7703 & 0.9174 & 0.9617 \\
     & OmniDepth ~\cite{zioulis2018omnidepth} & 0.2901 & 0.4838 & 0.7643 & 0.1450  & 0.6830 & 0.8794 & 0.9429 \\
     & Bifuse~\cite{BiFuse20} & 0.2048 & 0.3470 & 0.6259 & 0.1143 &  0.8452 & 0.9319 & 0.9632 \\
     & HoHoNet~\cite{SunSC21} & 0.1488 & 0.2862 & 0.5138 & 0.0871 &  0.8786 & 0.9519 & 0.9771 \\
     & SliceNet~\cite{Pintore2021Slicenet} & 0.1764 & 0.3296 & 0.6133 & 0.1045 & 0.8716 & 0.9483 & 0.9716 \\
     & Omnifusion~\cite{li2022omnifusion} & 0.1387 & 0.2724 & 0.5009 & 0.0893 & 0.8789 & 0.9617 & 0.9818 \\
     & Ours & \bf{0.1150} & \bf{0.2205} & \bf{0.4528} & \bf{0.0814} & \bf{0.9123} & \bf{0.9703} & \bf{0.9856} \\
     
    \bottomrule
  \end{tabular}
  \caption{Quantitative results on real-world indoor panorama depth estimation datasets-Stanford2D3D~\cite{armeni2017joint} and Matterport3D~\cite{Matterport3D}. Our model outperforms existing methods on all metrics.}
  \vspace{-1mm}
  \label{depth compare}
\end{table*}

\begin{figure*}
  \centering
  \includegraphics[width=1\linewidth]{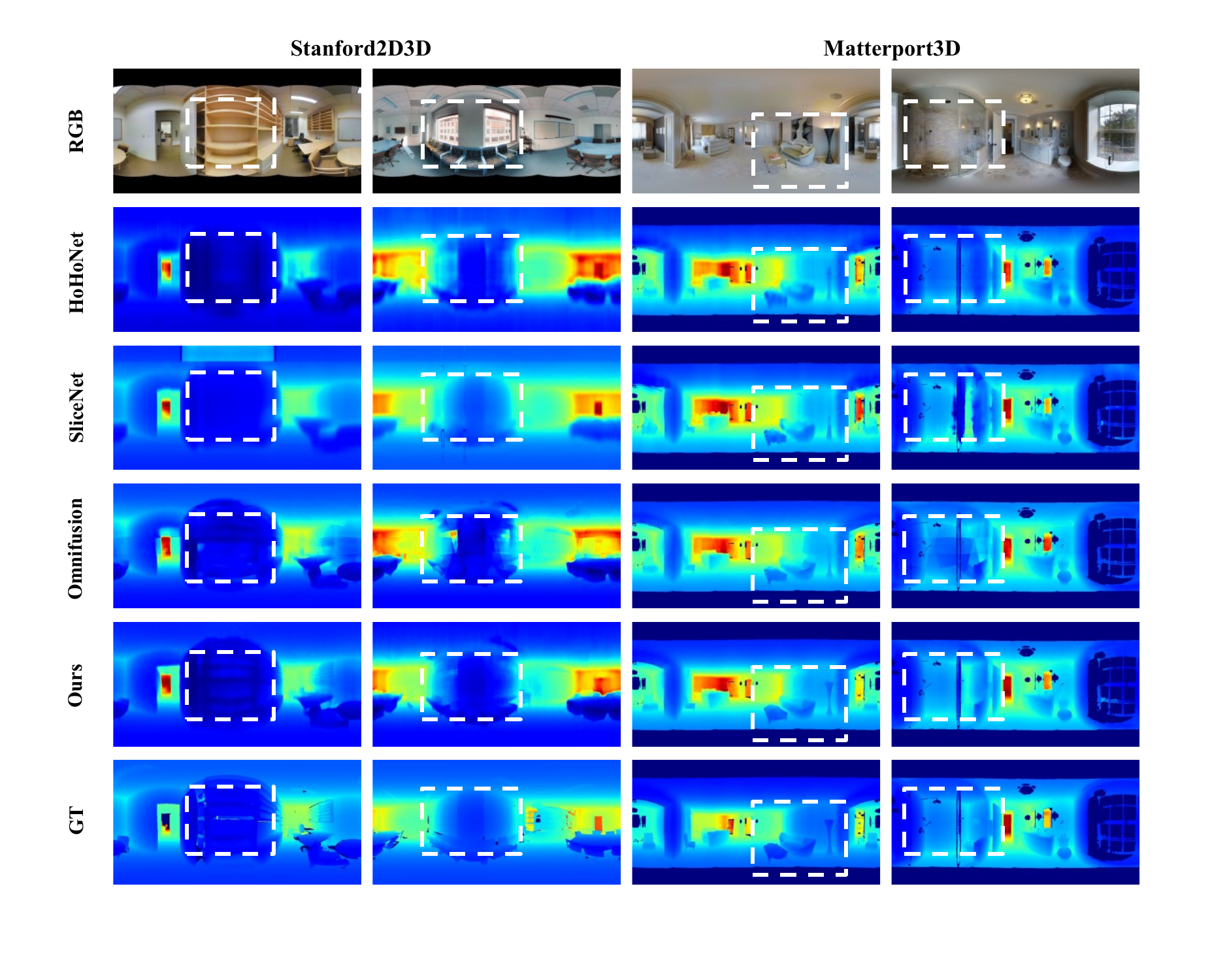}
  \vspace{-17mm}
  \caption{Qualitative results on Stanford2D3D~\cite{armeni2017joint} and Matterport3D~\cite{Matterport3D}. Our method generates sharp object edges and shows consistent indoor structure depth prediction in different scenes. The black spots stand for the invalid depth values.}
  \vspace{-3mm}
  \label{fig:quali}
\end{figure*}

\subsection{Datasets}

\noindent\textbf{Stanford2D3D~\cite{armeni2017joint}} is a real-world dataset consisting of 1,413 panoramas collected in 6 large-scale indoor areas. For depth estimation, we follow the official split and use area1, area2, area3, area4, area6 for training and area5 for testing. For semantic segmentation, we follow the previous works and use the official 3-fold split for training and evaluation. The resolution used for depth estimation is $512 \times 1024$ and $256 \times 512$ for semantic segmentation.

\noindent\textbf{PanoContext~\cite{zhang2014panocontext} and the extended Stanford2D3D~\cite{zou2018layoutnet}} are two cuboid room layout datasets. PanoContext~\cite{zhang2014panocontext} contains 514 annotated cuboid room layouts collected from SunCG~\cite{xiao2012recognizing} dataset. Zou et al.~\cite{zou2018layoutnet} collected 571 panoramas from Stanford2D3D~\cite{armeni2017joint} and annotated them with room layouts. The input resolution of both datasets is $512 \times 1024$. We follow the same splits of previous works~\cite{zou2018layoutnet, jiang2022lgt} for training and testing. 

\noindent\textbf{Matterport3D~\cite{Matterport3D}} is a large-scale RGB-D dataset that contains 10,800 panoramic images collected in 90 scenes. We use this dataset for our depth estimation experiment. We follow the official split that takes 7829 panoramas from 61 houses for training and the rest for testing. The resolution used in our experiment is $512 \times 1024$.


\subsection{Implementation details}
For depth estimation, we evaluate the performance of our model using the standard depth estimation metrics, including Mean Relative Error (MRE), Mean Absolute Error (MAE), Root Mean Square Error (RMSE), log-based Root Mean Square Error (RMSE(log)) and threshold-based precision $\delta^1$, $\delta^2$ and $\delta^3$. For semantic segmentation, we evaluate the performance using the standard semantic segmentation metrics class-wise mIoU and class-wise mAcc. For layout prediction, we use 3D Intersection over Union (3DIoU\%) to evaluate the performance.

We implement our model using Pytorch and train it on eight NVIDIA GTX 1080 Ti GPUs with a batch size of 16. We train the network using Adam optimizer, and the initial learning rate is set to 0.0001. For the depth estimation, we train our model on Stanford2D3D~\cite{armeni2017joint} for 100 epochs and Matterport3D~\cite{Matterport3D} for 60 epochs. We train our model 200 epochs on semantic segmentation datasets, and 1000 epochs on layout prediction datasets. We adopt random flipping, random horizontal rotation and random gamma augmentation for data augmentation. The default stride and interval for depth estimation are 32 and 128 while the stride is set to 16 for semantic segmentation.

\subsection{Evaluation on depth estimation datasets}
We evaluate our method against state-of-the-art panorama depth estimation algorithms in Table~\ref{depth compare}. The results are averaged by the best results from three training sessions. Note that the results of SliceNet~\cite{Pintore2021Slicenet} on Stanford2D3D~\cite{armeni2017joint} were reproduced by the fixed metrics and we retrain and re-evaluate a 2-iteration Omnifusion~\cite{li2022omnifusion} model on Matterport3D~\cite{Matterport3D} dataset. Our model outperforms existing models on all metrics on both datasets. Figure~\ref{fig:quali} shows the qualitative results of our model on Stanford2D3D~\cite{armeni2017joint} and Matterport3D~\cite{Matterport3D}. For the method that directly works on the panoramas~\cite{SunSC21, Pintore2021Slicenet}, they predict continuous background while lacking object details. Fusion-based method~\cite{li2022omnifusion} generates sharp depth boundaries while the strange artifacts caused by the patch-wise discrepancy lead to inconsistent depth prediction, e.g. the bookshelf in column 1 and the shower glass in column 4. It is not removable with its patch fusion module or iteration mechanism. With the help of our proposed Local2Global Transformer, our model preserves the geometric continuity of the room structure and shows superior performance even for some challenging scenarios, e.g. the windows in column 2. Our model also generates sharp object depth edges, e.g. the floor lamp and sofa in column 3. 


\subsection{Evaluation on semantic segmentation datasets}
\begin{figure}[t]
 \centering
 \includegraphics[trim={0.25cm 0.255cm 0.25cm 0.25cm},clip,width=0.5\textwidth]{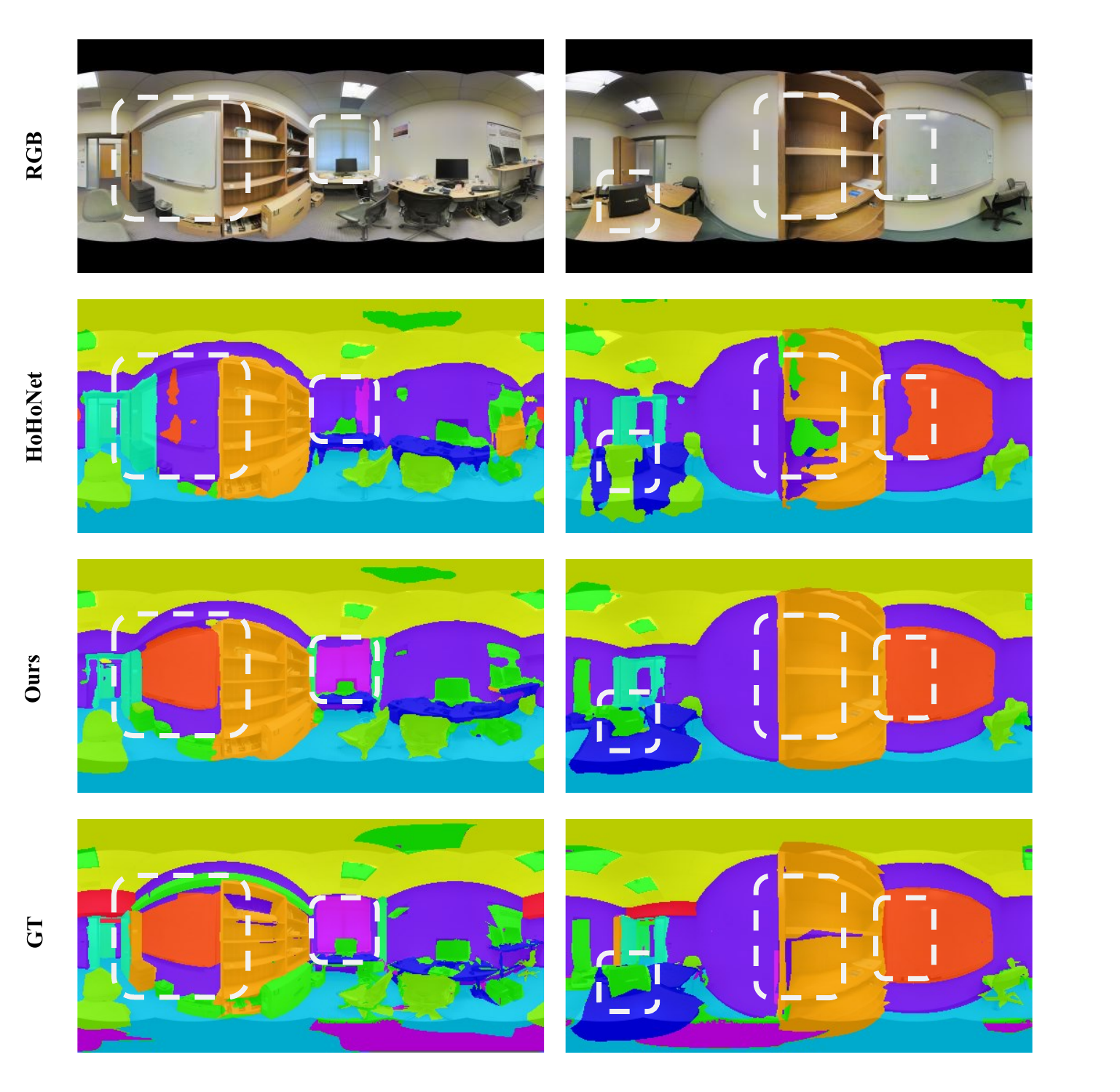}
 \vspace{-6mm}
 \caption{Qualitative results of semantic segmentation on Stanford2D3D~\cite{armeni2017joint} dataset.}
 \vspace{-6mm}
 \label{fig:semantic quali}
\end{figure}

We evaluate our method against state-of-the-art panorama semantic segmentation methods in Table~\ref{semantic quanti}. Our method improves the mIoU by 6.9\% and mAcc by 8.9\% against HoHoNet~\cite{SunSC21}. Note that we only use RGB panoramas as input. Figure~\ref{fig:semantic quali} shows the qualitative results of our semantic segmentation model on Stanford2D3D~\cite{armeni2017joint}. Our model shows a strong ability to segment out the objects with a smooth surface, e.g. the whiteboards and windows. The segmentation edges are natural and continuous. This is because the Local2Global Transformer successfully captures the geometric context of the object. Our model is also good at segmenting out the small objects from the background, e.g. the computer in both columns. Note that the segmentation boundaries of the ceiling and the walls generated by our model are very smooth against the previous work~\cite{SunSC21}, which shows the power of our panel geometry embedding network to learn the ERP distortion. Zoom in to view more details.


\begin{table}
  \centering
  \begin{tabular}{@{}lc|cc@{}}
    \toprule
     Method & Input & mIoU $\uparrow$ & mAcc $\uparrow$  \\
    \midrule
    TangentImg~\cite{eder2020tangent} & RGB-D & 41.8 & 54.9  \\
    HoHoNet~\cite{SunSC21} & RGB-D & 43.3 & 53.9  \\
    Ours & RGB & \bf{46.3} & \bf{58.7} \\
    \bottomrule
  \end{tabular}
  \caption{Quantitative results of semantic segmentation on Stanford2D3D~\cite{armeni2017joint} dataset.}
  \vspace{-3mm}
  \label{semantic quanti}
\end{table}

\subsection{Evaluation on layout estimation datasets}
We evaluate our method against state-of-the-art panorama layout estimation methods in Table~\ref{layout quanti}. 
By adding linear layers at the end of our depth estimation network, our model achieves competitive performance against state-of-the-art methods designed specifically for layout estimation.
Since our model is initially designed for dense prediction, it suffers an information loss in the process of upsampling and channel compression. 
Our layout prediction model shares the same structure with the depth estimation model before the linear layers. We can activate this model with the weights pretrained on depth estimation datasets to reduce the training overhead. 
We find that our layout prediction model has the best performance when the stride is 64 and the interval is 128 so we use this setup for experiments on both datasets.

\begin{table}
  \centering
  \begin{tabular}{@{}l|cc@{}}
    \toprule
     Method & PanoContext & Stanford2D3D  \\
    \midrule
    LayoutNet v2~\cite{zou2021manhattan} & 85.02 & 82.66  \\
    DuLa-Net v2~\cite{zou2021manhattan} & 83.77 & \bf{86.60}  \\
    HorizonNet~\cite{SunHorizon19} & 84.23 & 83.51 \\
    AtlantaNet~\cite{pintore2020atlantanet} & - & 83.94 \\
    LGT-Net~\cite{jiang2022lgt} & \bf{85.16} & 85.76\\
    Ours & 84.52 & 85.91\\
    \bottomrule
  \end{tabular}
  \caption{Quantitative results of layout estimation on PanoContext dataset and Stanford2D3D~\cite{armeni2017joint} dataset in 3DIoU (\%). Following the previous works~\cite{zou2018layoutnet, jiang2022lgt}, we use the combination of PanoContext~\cite{zhang2014panocontext} and Stanford2D3D~\cite{zou2018layoutnet} for training.}
  \vspace{-1mm}
  \label{layout quanti}
\end{table}


\begin{table*}[!t]
  \centering
  \begin{tabular}{@{}l|c|ccccccc@{}}
    \toprule
    Method & Train Mem.& MRE &MAE & RMSE & $\delta^1$ & $\delta^2$ \\
    \midrule
    \multirow{1}{*}{}Baseline & 10231 & 0.1033 & 0.1859 & 0.3212 & 0.8976 & 0.9741 \\
     \midrule
     \multirow{2}{*}{}Baseline + Geo(G) & 10371 & 0.1029 & 0.1861 & 0.3205 & 0.8980 & 0.9765 \\
     Baseline + Geo(G+L) & 10509 &  0.1000 & 0.1815 & 0.3149 & 0.9012 & 0.9775 \\
     \midrule
     \multirow{2}{*}{}Baseline + Transformer(P) & 10359 & 0.0904 & 0.1652& 0.3058 & 0.9123 & 0.9776 \\
     Baseline + Transformer(P+W) & 10379 &  0.0854 & 0.1610 & 0.3016 & 0.9164 & 0.9785 \\
     \midrule
     \multirow{2}{*}{}Baseline + Geo(G+L) + Transformer(P) & 10639 &  0.0851 & 0.1572 & 0.2954 & 0.9218 & 0.9789 \\
     Baseline + Geo(G+L) + Transformer(P+W) & 10659 & \bf{0.0829} & \bf{0.1520} & \bf{0.2933}  & \bf{0.9242} & \bf{0.9796}\\     
     
    \bottomrule
  \end{tabular}
  \caption{Ablation study about the impact of each PanelNet component. "P" and "W" stands for the Panel Blocks and Window Blocks in Local2Global Transformer. "G" and "L" stands for the global and local feature embedded by the panel geometry embedding network. "Train Mem." stands for the GPU memory (MB) overhead of training our model on a single GTX-1080Ti GPU, the batch size is 2. }
  \vspace{-2mm}
  \label{Ablation conponent}
\end{table*}

\subsection{Ablation study}
\label{subSection: Ablation study}
In this section, we conduct ablation studies to evaluate the impact of the elements and hyper-parameters of our model on Stanford2D3D~\cite{armeni2017joint} dataset for depth estimation.

\begin{figure}
  \centering
  \includegraphics[width=1\linewidth]{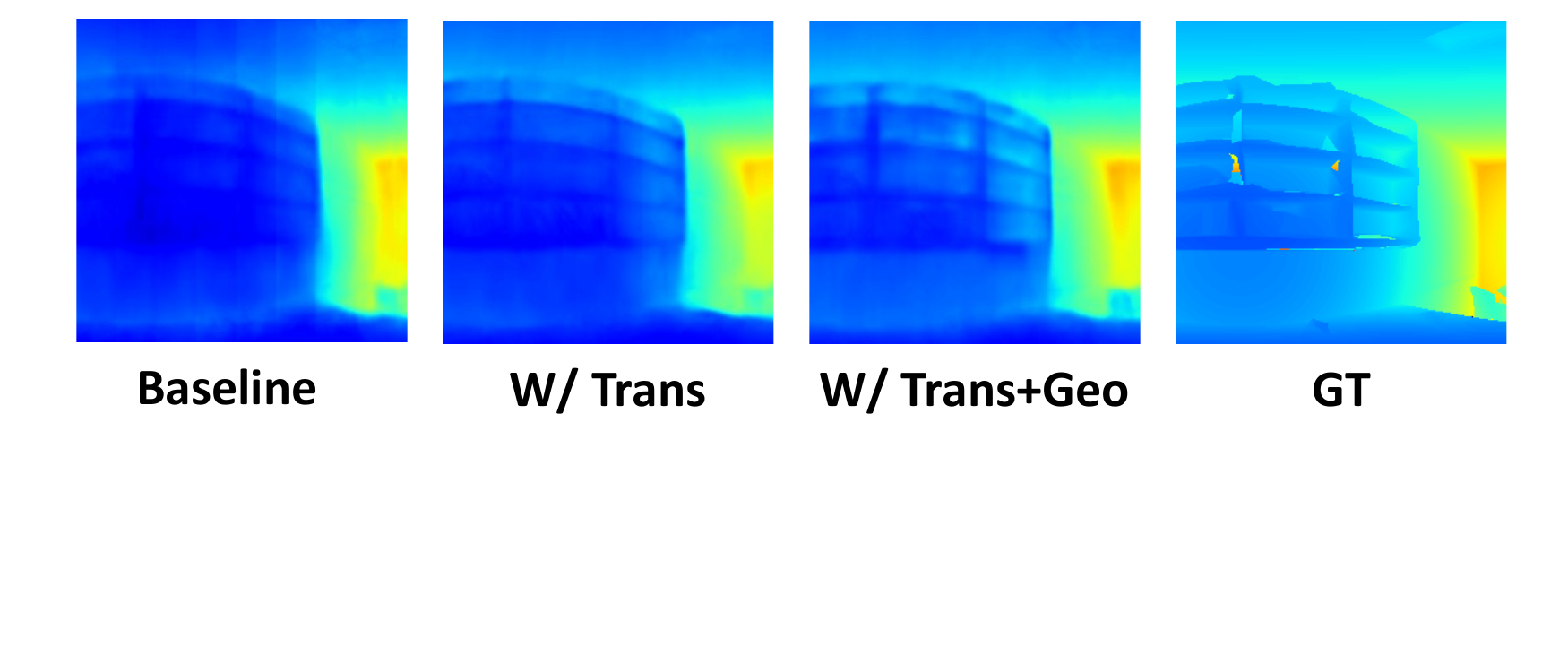}
  \vspace{-19mm}
  \caption{Qualitative effects of each network element. The baseline model is the pure CNN model listed in the first row of Table~\ref{Ablation conponent}.}
  \label{fig:ablation_quali}
  \vspace{-2mm}
\end{figure}

\noindent\textbf{Effects of individual network components } We conduct an ablation study to evaluate the impact of each component in our model, presented in Table~\ref{Ablation conponent}. The stride is set to 32 and the interval is 128 for all networks. We conduct our baseline model with a ResNet-34~\cite{he2016deep} encoder and a depth decoder as illustrated in Section~\ref{Network architecture}. Since partitioning the entire panorama into panels with overlaps greatly increase the computational complexity, we use ResNet-34 rather than vision Transformers as backbones. 
As we observed in Table~\ref{Ablation conponent}, the performance improvement of adding the panel geometry embedding network to the pure CNN structure is small since the network's ability to aggregate distortion information with image features is low. 
By applying the Local2Global Transformer as a feature processor, our baseline network gains a significant performance improvement on all evaluation metrics. Benefiting from the information aggregation ability of our proposed Local2Global Transformer, the panel geometry embedding network fully performs its ability on distortion perception and improves the performance both quantitatively and qualitatively. As shown in Figure~\ref{fig:ablation_quali}, the combination of Local2Global Transformer and panel geometry embedding network leads to the clearest object edges. 
We further test panel-wise relative position embedding similar to LGT-Net~\cite{jiang2022lgt} for Panel Blocks. However, it brings minimal performance improvements on depth estimation while increasing the computational complexity. 

\begin{table}
\centering
\scalebox{0.8}{

  \begin{tabular}{@{}l|ccccc@{}}
    \toprule
    Method & MRE & MAE& RMSE & $\delta^1$ & $\delta^2$ \\
    \midrule
    Omnifusion w/o Trans. & 0.1132 & 0.1932  & 0.3248 & 0.8728 & 0.9690 \\
    PanelNet w/o Trans. & 0.1000 & 0.1815 & 0.3149 & 0.9012 & 0.9775 \\
    Omnifusion w/ L2G & 0.1054 & 0.1918 & 0.3351 & 0.8870 & 0.9754 \\
    PanelNet w/ L2G & 0.0829 & 0.1520 & 0.2933 & 0.9242 & 0.9796 \\
    \bottomrule
  \end{tabular}}
  \vspace{-2mm}
    \caption{Ablation study on the usefulness of panel representation against tangent images. }
  \label{tab:effectiveness of panel}
  \vspace{-5mm}
\end{table}

We conduct an ablation study to further validate the usefulness of panel representation against tangent images. We use Omnifusion~\cite{li2022omnifusion} as a comparison since it has a similar input format and can be trained via the same encoder-decoder CNN architecture with our model. As shown in Table~\ref{tab:effectiveness of panel}, the panel representation with pure CNN architecture outperforms the original Omnifusion~\cite{li2022omnifusion}, which demonstrates the superiority of panel representation. We replace the default transformer of Omnifusion~\cite{li2022omnifusion} with Local2Global Transformer. However, the Local2Global Transformer doesn't bring a huge performance improvement for tangent images since the discontinuous tangent patches lower the ability of the Window Blocks to aggregate local information in the vertical direction which reduces the continuity of depth estimation for gravity-aligned objects and scenes. On the contrary, the vertical continuity is preserved within the vertical panels. With the panel representation, the Local2Global Transformer exerts its greatest information aggregation ability. 

\noindent\textbf{Effects of panel size and stride } We further study the effect of panel size on the performance and speed, as displayed in Table~\ref{tab:panel size}. The FPSs are obtained by measuring the average inference time on a single NVIDIA GTX 1080Ti GPU. We observe that for the models that have the same panel interval, i.e. width, a smaller stride enhances the performance. For the same stride, the models with larger panels have better performance. Theoretically, smaller strides improve performance because horizontal consistency is preserved by the more overlapping area of consecutive panels. Larger panels also lead to better performance because larger panels provide larger FoV, which contains more geometric context within a panel. However, we observe that keep increasing the interval may have a negative impact on performance. The larger panel brings higher computational complexity, which forces the stride to increase to reduce the computational overhead. This makes the performance gain brought by the larger FoV be wiped out by the consistency loss due to fewer overlaps. 
To gain the best performance, we set the interval to 128 and the stride to 32 for most of our experiments.

\section{Conclusion}
\label{sec:conclusion}

\begin{table}
  \centering
  \begin{tabular}{@{}ccc|c|ccc@{}}
    \toprule
    I & S & \#Panel& FPS &
    MRE & RMSE & $\delta^1$ \\
    \midrule
    64 & 16 & 128 & 6.4 & 0.0866 & 0.3040 & 0.9181 \\
    64 & 32 & 64 & 12.4 & 0.0909 & 0.3207 & 0.9102\\
    64 & 64 & 32 & 24.4 & 0.0952 & 0.3319 & 0.9041\\
    128 & 32 & 32 & 6.9 & \bf{0.0829} & \bf{0.2933} & \bf{0.9242}\\
    128 & 64 & 16 & 13.5 & 0.0892 & 0.3109 & 0.9172\\
    128 & 128 & 8 & 25.7 &0.0920 & 0.3181 & 0.9103\\
    256 & 64 & 16 & 7.5 & 0.0894 & 0.3047 & 0.9132\\
    256 & 128 & 8 & 13.9 & 0.0908 & 0.3069 & 0.9182\\
    256 & 256 & 4 & \bf{26.4} & 0.0986 & 0.3248 & 0.8991\\
    \bottomrule
  \end{tabular}
  \caption{Ablation study on the impact of panel size and stride. "I" and "S" stand for the panel interval and stride mentioned in Section~\ref{Network architecture}. "\#Panel" stands for the number of panels.}
  \label{tab:panel size}
  \vspace{-3mm}
\end{table}

We present PanelNet, a framework that understands indoor environments from 360 images. Based on the essential properties of indoor equirectangular projection (ERP), we introduce a novel panel representation to model the indoor scene. We design a panel geometry embedding network to encode both local and global geometric features which reduces the negative impact of ERP distortion implicitly. We introduce Local2Global Transformer for information aggregation, which greatly improves the performance of our model by successfully aggregating the local information within a panel and capturing panel-wise global context. Our model outperforms existing panorama depth estimation approaches on all evaluation metrics and achieves competitive results on 360 indoor semantic segmentation and layout estimation.

{\small
\bibliographystyle{ieee_fullname}
\bibliography{11_references}
}


\end{document}